\documentclass{article}

\usepackage{microtype}
\usepackage{multicol}
\usepackage{graphicx}
\usepackage{subcaption}
\usepackage{booktabs} %

\usepackage{hyperref}

\usepackage[accepted]{icml2024}

\usepackage{amsmath}
\usepackage{amssymb}
\usepackage{mathtools}
\usepackage{amsthm}
\usepackage{enumitem}
\usepackage{lipsum} 

\usepackage[capitalize,noabbrev]{cleveref}

\theoremstyle{plain}

\theoremstyle{definition}

\theoremstyle{remark}

\usepackage[textsize=tiny]{todonotes}

\newcommand{\xxnote}[3]{}
\renewcommand{\xxnote}[3]{\color{#2}{#1: #3}}

\newcommand{\method}{\textsc{ASMR}}
\usepackage{xspace}
\newcommand{\ksp}{\textit{k}-space\xspace}

\icmltitlerunning{Adaptive Sampling of k-Space in Magnetic Resonance for Rapid Pathology Prediction}

\begin{document}

\twocolumn[
\icmltitle{Adaptive Sampling of $k$-space in Magnetic Resonance \\ for Rapid Pathology Prediction}

\icmlsetsymbol{equal}{*}

\begin{icmlauthorlist}
\icmlauthor{Chen-Yu Yen}{equal,yyy}
\icmlauthor{Raghav Singhal}{equal,yyy}
\icmlauthor{Umang Sharma}{yyy}
\icmlauthor{Rajesh Ranganath}{yyy}
\icmlauthor{Sumit Chopra}{yyy}
\icmlauthor{Lerrel Pinto}{yyy}

\end{icmlauthorlist}

\icmlaffiliation{yyy}{New York University}

\icmlcorrespondingauthor{Chen-Yu Yen}{chenyu.yen@nyu.edu}
\icmlcorrespondingauthor{Raghav Singhal}{rsinghal@nyu.edu}

\icmlkeywords{Reinforcement Learning, Healthcare, MRI, Machine Learning, ICML}

\vskip 0.3in
]

\printAffiliationsAndNotice{\icmlEqualContribution} %

\begin{abstract}
Magnetic Resonance (MR) imaging, despite its proven diagnostic utility, remains an inaccessible imaging modality for disease surveillance at the population level. A major factor rendering MR inaccessible is lengthy scan times. An MR scanner collects measurements associated with the underlying anatomy in the Fourier space, also known as the \textit{k}-space. Creating a high-fidelity image requires collecting large quantities of such measurements, increasing the scan time. Traditionally to accelerate an MR scan, image reconstruction from under-sampled \ksp data is the method of choice. However, recent works show the feasibility of bypassing image reconstruction and directly learning to detect disease directly from a sparser learned subset of the \textit{k}-space measurements. In this work, we propose  Adaptive Sampling for MR (ASMR), a sampling method that learns an adaptive policy to sequentially select \textit{k}-space samples to optimize for target disease detection. On $6$ out of $8$ pathology classification tasks spanning the Knee, Brain, and Prostate MR scans, ASMR reaches within $2\%$ of the performance of a fully sampled classifier while using only $8\%$ of the \ksp, as well as outperforming prior state-of-the-art work in \textit{k}-space sampling such as EMRT, LOUPE, and DPS.
\end{abstract}

\section{Introduction}
\label{sec:intro}

Magnetic Resonance (MR) imaging is a diagnostic medical imaging tool that exhibits superior soft-tissue contrast in its generated images. 
This has led to MR scans being the gold standard for diagnosing a range of cancers \citep{winawer1997colorectal,ilic2018prostate,elmore1998ten}, and musculoskeletal disorders \citep{dean2011role,elvenes2000magnetic}. 
An MR machine sequentially measures the responses to radio-frequency pulses of a human body placed in a magnetic field to generate cross-sectional images (slices) of the anatomy \citep{zhi2000principles,zbontar2018fastmri}. The measurements are acquired in the Fourier space, also known as the \textit{k}-space \citep{zhi2000principles} in the MR community. The spatial image is then constructed by applying a multi-dimensional Fourier transform to the \textit{k}-space data. High-resolution MR images require a large number of \textit{k}-space measurements, leading to increased scan times (up to $40$ minutes, depending on the anatomy) \citep{zbontar2018fastmri}.

\begin{figure}[t]
    \centering
    \includegraphics[width=\linewidth]{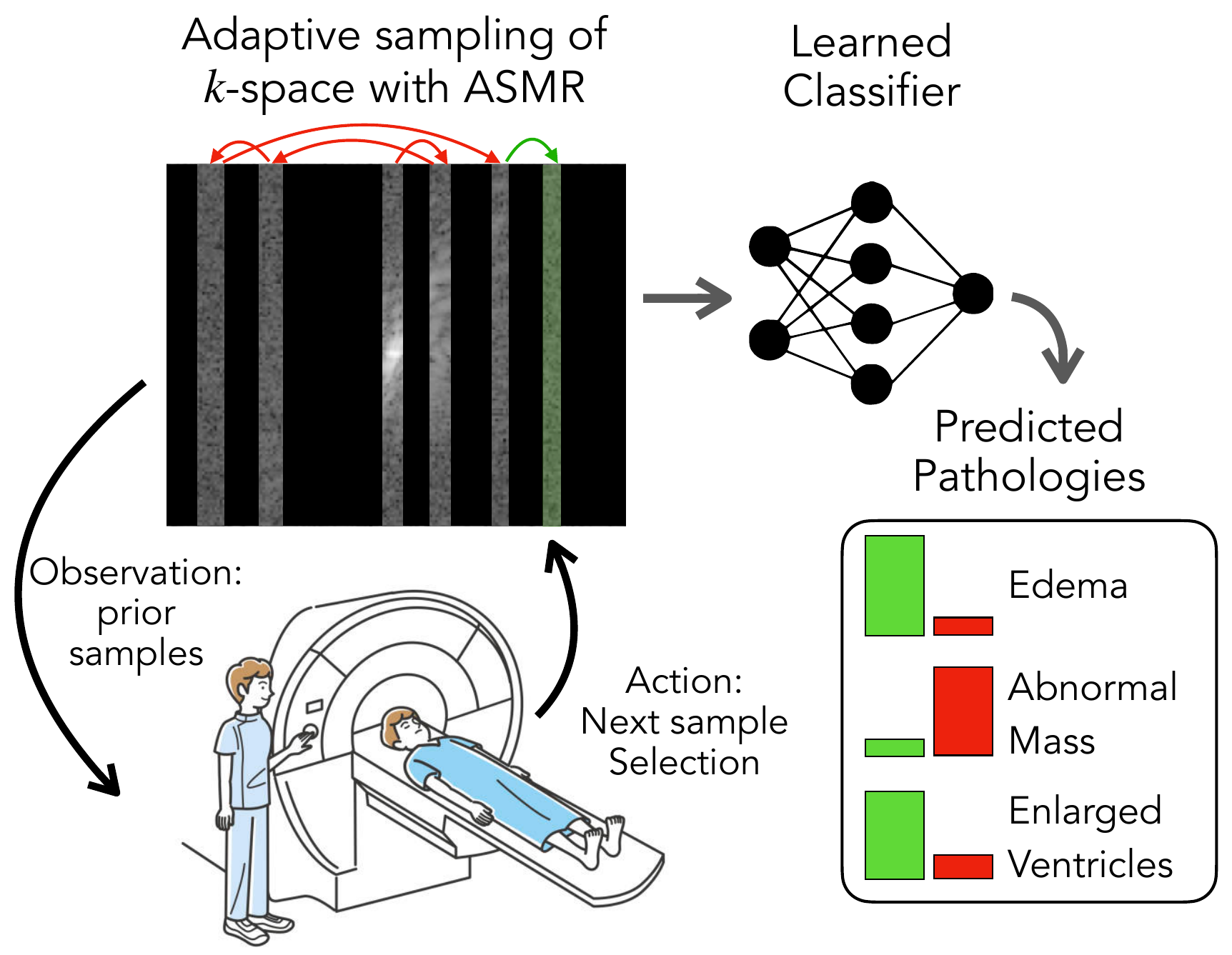}
    \caption{\textbf{Overview of \method{}.} Given a set of prior samples from the MR scanner, \method{} proposes the next sample to collect.These sampling steps are repeated for a fixed set of iterations $T$, at which point the collected samples are used by a classifier to predict the presence of a fixed set of pathologies. \method{} is trained to optimize classification performance and sidesteps reconstruction altogether.}
    \label{fig:introduction}
\end{figure}

Reducing MR scan times is an active area of research. A common approach is to under-sample the \textit{k}-space and then use image reconstruction techniques such as compressed sensing \citep{lustig2008compressed,zbontar2018fastmri,donoho2006compressed} and deep learning \citep{sriram2020end,chung2022score} to reconstruct the underlying image accurately. 
A recent challenge organized by NYU Langone \citep{knoll2020advancing} shows that at a $12.5\%$ sampling rate, reconstructed images started missing pathologies. This is understandable as reconstruction metrics, such as SSIM \citep{wang2004image}, measure the similarity between moments of localized patches rather than optimizing for diagnostic performance. To overcome this, recent work \citep{singhal2023feasibility,tolpadi2023k2s} shows that we can directly detect the presence of a pathology and bypass reconstruction altogether. Furthermore, \citet{singhal2023feasibility} show that directly classifying from a sparse subset of the \textit{k}-space yields similar performance to using the entire \textit{k}-space. However, both \citet{singhal2023feasibility,tolpadi2023k2s} use a heuristic approach to find the sampling pattern. 

Typically, the sampling patterns are heuristic or hand-crafted \citep{lustig2007sparse}, based on the assumption that natural images have rapidly decaying Fourier coefficients \citep{lustig2007sparse}. However, \citet{bahadir2019learning,jin2019self,pineda2020active,huijben2019deep} show that learning the sampling pattern to optimize for the underlying task offers significant improvements over heuristic sampling patterns. In practice, a heuristic sampling pattern, such as an equi-spaced pattern \citep{johnson2022deep}, at a fixed sampling rate is used for each patient \citep{zbontar2018fastmri}. However, a heuristic pattern cannot adapt to the collection of previously acquired $k$-space samples.

In this work, we introduce Adaptive Sampling for Magnetic Resonance (\method{}). Using the fact that the \textit{k}-space data collection is a sequential process, we pose the problem of learning the sampling pattern as a decision-making problem, which can then be optimized using reinforcement learning (RL). Concretely, at every sampling time step, the collection of previous samples serves as the observation, the choice of the next sample serves as the action, and the reward is computed by measuring the log-likelihood between the labels and the sub-sampled $k$-space. 
See \cref{fig:introduction} for an overview of \method{}. The policy for selecting actions is optimized using Proximal Policy Optimization. Unlike prior work that uses RL for reconstruction \citep{bakker2020experimental,jin2019self,pineda2020active}, \method{} is trained to maximize the classification performance.

To understand the performance of \method{}, we run experiments on the \textsc{FastMRI} dataset \citep{mathieu2013fast,zhao2021fastmri+}, consisting of volumetric brain, knee and abdominal scans. For simplicity, we use Cartesian sampling \citep{zhi2000principles,lustig2008compressed} that selects columns of the $2$D complex-valued \textit{k}-space matrix. Our experiments reveal the following findings:
\begin{enumerate}
    \item \method{} outperforms state-of-the-art non-adaptive sampling patterns such as EMRT \citep{singhal2023feasibility} on 6 out of 8 classification tasks. Compared to learned probabilistic sampling patterns like LOUPE \citep{bahadir2019learning} and DPS, \method{} achieves an improvement of atleast 2.5\% in the AUC metrics on 7 out of 8 tasks (See \cref{main_all}).
   
    \item \method{} outperforms adaptive sampling methods \citep{bakker2020experimental} that are optimized for reconstruction, providing an absolute gain of 2.88\% and 7\% in the AUC metrics on the Knee and Brain datasets respectively (See \cref{fig:recon_brain}).

    \item Compared to an image-based classifier that uses the entire \ksp data as input, \method{} comes within $2\%$ of its performance on 6 out of 8 tasks with a $12.5\times$ acceleration factor (See \cref{main_all}).
\end{enumerate}

Our code can be found at \href{https://adaptive-sampling-mr.github.io/}{adaptive-sampling-mr.github.io}.

\section{Related Work}\label{sec:related_work}

\method{} builds upon several prior works in \ksp-based prediction, mask sampling strategies, and adaptive learning.

\paragraph{Pathology prediction from sparse \ksp:} \citet{singhal2023feasibility} show that directly classifying from sparsely acquired %
\ksp data yields classification performance similar to an image-based model that uses as input images reconstructed using the entire \ksp data. The sampling pattern they use to acquire the sparse \ksp data is optimized using a non-adaptive variable density sampling prior \citep{lustig2007sparse}.
Similarly, \citet{tolpadi2023k2s} show that directly generating segmentations as the main objective with a secondary reconstruction objective can also accelerate pathology prediction without any significant drop in performance. Interestingly, \citet{tolpadi2023k2s} show that there was no correlation between segmentation and reconstruction performance. In both works, the sampling patterns used were heuristic-based, non-adaptive, and not optimized for the downstream task.

\paragraph{Learned non-adaptive sampling methods:}
Several prior works \cite{bahadir2019learning,huijben2019deep,weiss2020joint} have looked at learning a probabilistic model over sampling patterns jointly with a reconstruction model. \citet{gozcu2019rethinking,weiss2019pilot} learn a fixed sampling pattern using a pre-trained reconstruction model and the reconstruction error as a scoring rule for sampling patterns. 
The sampling patterns learned in these methods are non-adaptive and only produce sampling patterns for a fixed sampling budget. In \cref{fig:learning-based-baseline} we compare ASMR against sampling patterns optimized for classification using \textsc{loupe} \citep{bahadir2019learning} and \textsc{dps} \citep{huijben2019deep}.

\paragraph{Learned adaptive sampling methods:} More recently,
\citet{bakker2020experimental,pineda2020active,zhang2019reducing,jin2019self} pose learning an adaptive sampling pattern as an \textsc{rl} problem. For example \citet{bakker2020experimental} first trains a reconstruction model and then uses the reconstruction loss as a reward to train the policy. \citet{zhang2019reducing} also presents an active acquisition method, where instead of learning a policy model, they jointly train a reconstruction model and evaluator model. Similarly, \citet{jin2019self} jointly learn a reconstruction model and policy network using Monte-Carlo tree search, similar to AlphaGo \citep{silver2016mastering}. All these works use reconstruction metrics as the reward, which is often sub-optimal for pathology classification \citep{singhal2023feasibility}. \Cref{fig:recon_brain} shows that \method{}, which is directly optimized for classification improves upon reconstruction-based sampling.

\paragraph{Reinforcement learning for adaptive sampling:} 
Reinforcement learning for adaptive sampling is a promising area of research in the fields of gaming~\cite{silver2016mastering}, finance~\cite{rl_finance}, and chatbots~\cite{achiam2023gpt}. For instance, in the robotics domain, it has been used to improve controllers~\cite{zucker2008adaptive,pan2022marlas}, and path planning~\citep{pan2022marlas}. 
Additionally, it has applications in autonomous balloon navigation~\cite{bellemare2020autonomous}, chip design and hardware optimization~\cite{kumar2022datadriven, chen2023adatest}. However, unlike applications where a large amount of data is available or new data can be easily synthesized, one of our challenges is the limited dataset size. Unlike scenarios where the reward is well-defined, another challenge is the choice of reward suitable for fast pathology classification.

\section{Background and Setup}
In this section, we provide background for understanding \method{}, which includes understanding how an MR scanner works, details on \ksp classification, and adaptive sampling with reinforcement learning.

\begin{figure}[t!]
 \centering
 \includegraphics[width=\linewidth]{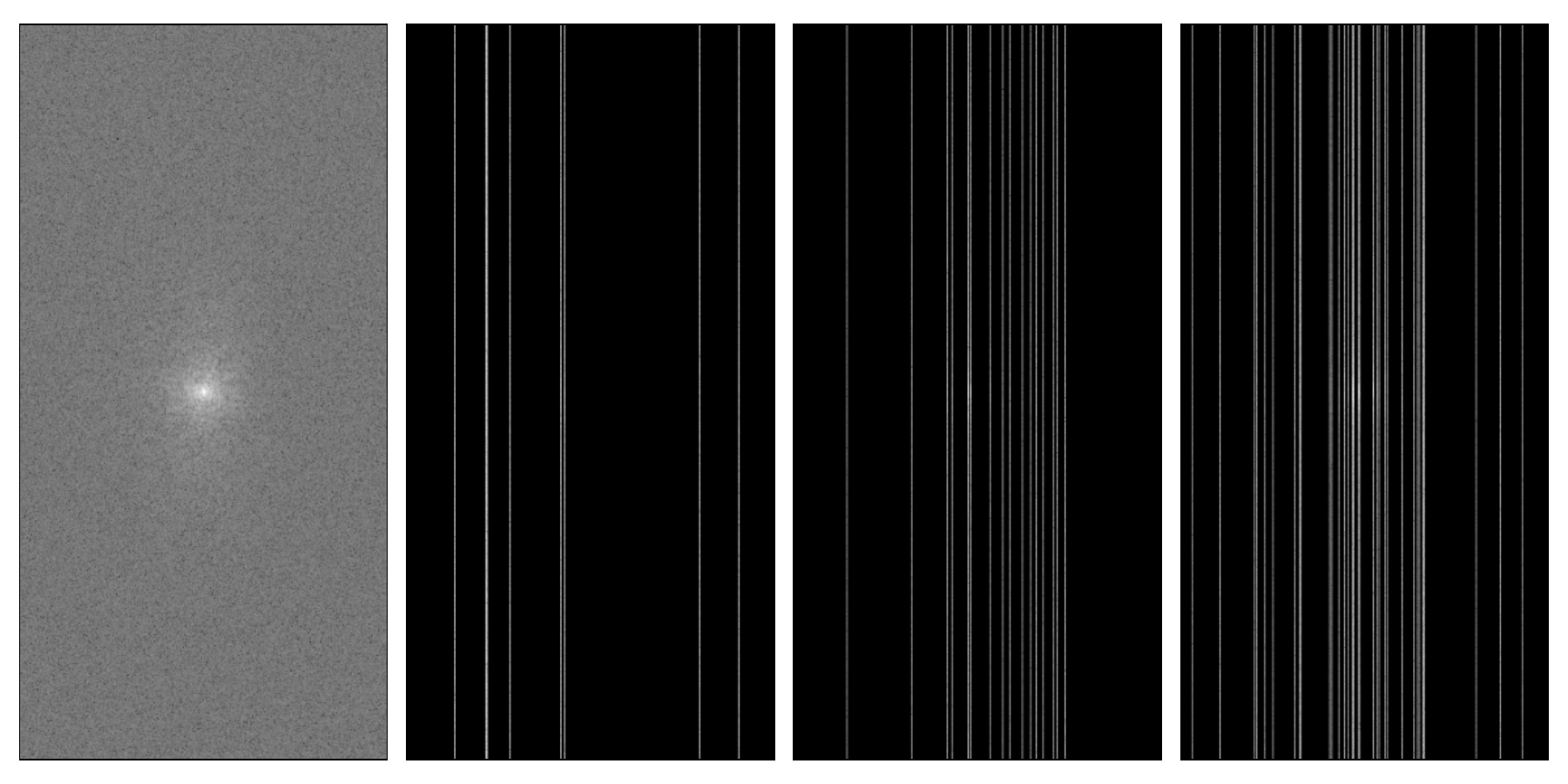}
 \caption{Cartesian Sampling on \ksp at different sampling rates. The leftmost image shows the complete \ksp followed 2.5\%, 5\% and 10\% sampling rates (from left to right)}
 \label{fig:undersampled_ksp}
\end{figure}

\subsection{Magnetic Resonance (MR) Environment} 
\label{sec:env}
MR imaging functions indirectly, acquiring frequency and phase measurements of the anatomy being imaged instead of direct spatial data. A receiver coil, situated near the target area,
acquires a collection of measurements encoding the human body's response to a combination of applied electromagnetic fields, including a static spatially homogeneous magnetic field, spatially varying magnetic fields (gradients), and radio-frequency pulses (the observed measurements) \citep{zhi2000principles}. These measurements of the body's response to different radio-frequency pulses correspond to points in the \ksp and are stored in a matrix form called the \ksp matrix. A \ksp matrix is a complex-valued matrix $\mathbf{x} \in \mathbb{C}^{d_r \times d_c}$, which is the Fourier Transform of the underlying $2$D MR image. Hence, an image can be reconstructed using an Inverse Fourier Transform of the \ksp. 

We represent the \ksp sampling pattern using a two-dimensional binary mask $\mathbf{s} \in \prod_{i, j = 1}^{d_r, d_c} \{0, 1\}$, such that $\mathbf{s}_{i, j} = 1$ indicates that $\mathbf{x}_{i, j}$ has been sampled. Typically, the \ksp is retrospectively under-sampled by multiplying the full \ksp $\mathbf{x}$ with the binary mask $\mathbf{s}$ \citep{zbontar2018fastmri}, which we denote as $\mathbf{x}_\mathbf{s} = \mathbf{x} \odot \mathbf{s}$, where $\odot$ stands for element-wise multiplication.

The \ksp measurements are acquired \textit{sequentially}. A MR scanner has physical restraints on how it can sample. Some examples of physically feasible trajectories include Cartesian, spiral, and radial \cite{zbontar2018fastmri,weiss2019pilot}. In this work, we consider Cartesian sampling, which involves sampling columns of the \ksp matrix sequentially. See \cref{fig:undersampled_ksp} for examples of Cartesian sampling on \ksp. It can take a long time to acquire measurements sufficient for reconstructing a diagnostic quality image, thereby increasing the cost of the scan and making them infeasible for population-level screenings. 

In the next section, we discuss recent work that bypasses image reconstruction and directly predicts the presence or absence of a set of pathologies. 

\subsection{Classification from \ksp data}\label{sec:classification_background}
\label{inferece-time}

Given a labeled dataset $(\{(\mathbf{x}_i, \mathbf{y}_i)\}_{i=1}^N)$ consisting of \ksp data $\mathbf{x}$ and pathology labels $\mathbf{y}$, we follow \citet{singhal2023feasibility} and directly train a classifier from the retrospectively under-sampled \ksp $\mathbf{x}_\mathbf{s}$ using the \textsc{kspace-net} classifier $q_\phi(\mathbf{y} \mid \mathbf{x}_\mathbf{s})$. Importantly, the \textsc{kspace-net} classifier $q_\phi$ takes the complex-valued \ksp as input and does not rely on reconstructed images. 
Similar to \citet{jethani2021have,jethani2023don} and \citet{singhal2023feasibility} we train the \textsc{kspace-net} classifier using randomly masked out \ksp data, where the masks are drawn independently from the variable density prior \citet{lustig2007sparse}. Models trained in such a way can then be used to score masks. For instance, we can estimate the mutual information \citep{cover1999elements} between $\mathbf{x}_ \mathbf{s}$ and the label  $\mathbf{y}$, upto a constant, using the classifier $q_\theta$:
\begin{align}\label{eq:scoring_rule}
    \mathbf{V}_\theta(\mathbf{s}) = \mathbb{E}_{q(\mathbf{x}_{\mathbf{s}})q(\mathbf{y} \mid \mathbf{x}_{\mathbf{s}})} \log {q_\theta(\mathbf{y} \mid \mathbf{x}_{\mathbf{s}})}
\end{align}
In the \cref{sec:method}, we show how this classifier can be used to design rewards for the policy. 
\subsection{Reinforcement Learning}
\label{sec:rl_background}

The reinforcement learning problem formulates the decision-making problem as Markov Decision Process (MDP)
$(\mathcal{S}, \mathcal{A}, p_0(s), p(s'|s, a), R(s, a), \gamma, T)$, where $a$ is an action, $s$ is the state, \(\mathcal{S}\) is a state space, \(\mathcal{A}\) is an action space, \(p_0(s)\) is a distribution of initial states, \(p(s'|s, a)\) is the environment dynamics, \(R(s, a)\) is a reward function, \(\gamma\) is a discount factor, and $T$ is the task horizon. The agent interacts with the MDP according to a policy \(\pi(a|s)\). 
The goal of reinforcement learning is to obtain a policy that maximizes the cumulative discounted returns:
$ \pi^* = \arg\max_{\pi} \mathbb{E}_\pi \left[ \sum_{t=0}^{T} \gamma^t R(s_t, a_t)  \right]$. For more details on RL we refer the reader to \citet{sutton2018reinforcement}.

\paragraph{Proximal Policy Optimization (PPO):}
PPO \citep{ppo} is an RL algorithm that learns a policy $\pi_\theta$
and a value function $V_\phi$ with the goal of finding an optimal policy for an MDP. An actor proposes an action to the environment, and a critic estimates a value for the action proposed by the actor.
To this end, PPO optimizes a surrogate objective 
\begin{equation}
    L = \mathbb{E}\left[\min(l_t(\theta) \hat{A}_t, \text{clip}(l_t(\theta), 1-\varepsilon, 1+\varepsilon) \hat{A}_t)\right]
    \label{ppo}
\end{equation}
where $l_t(\theta) = \frac{\pi_\theta(a_t | s_t)}{\pi_{\text{old}}(a_t | s_t)}
$ denotes the likelihood ratio between the new and old policies, and  
$\hat{A}_t = R_t + \gamma V_\phi(s_{t+1}) - V_\phi(s_t)$ denotes an advantage function at timestep $t$. During training, PPO penalizes large changes to the policy to improve the stability of learning. For this work, we build our agent on top of an open-source implementation of PPO in~\cite{cleanrl}.

\section{Adaptive Sampling for Magnetic Resonance}
\label{sec:method}
In this section, we introduce Adaptive Sampling for Magnetic Resonance (\method{}). \method{} learns an adaptive sampling method for direct \ksp classification by training an RL agent to select \ksp samples sequentially. 
Unlike standard RL settings with access to an online environment, we retrospectively under-sample the \ksp by applying a binary mask $\mathbf{s}$ to the fully-sampled \ksp $\mathbf{x}$. Furthermore, the action space and reward function require additional consideration given the constraints of the MR pathology prediction problem, such as data imbalance and the input being in the Fourier domain. 

\subsection{Formulating \ksp selection as an RL Problem} 
\label{sec:prob-formulation}

To frame the selection of \ksp measurements as a RL problem, we simulate environments using a dataset $\mathcal{D} = \{(\mathbf{x}_i,\mathbf{y}_i)\}_{i=1}^n$, where each \ksp sample $\mathbf{x}_i \in \mathbb{C}^{d_r\times d_c}$ represents one environment and has an associated pathology label $\mathbf{y}_i$. We employ a Cartesian sampling pattern, where an acquisition corresponds to a column of the \ksp matrix. 

Within each environment, the goal of the RL agent is to sequentially select a set of  $T = |d_c| \times \alpha$ \ksp columns to maximize classification performance, where $T$ represents the sampling budget and serves as our episode length.
The agent is tasked to take action $a_t \in \{0, 1, \dots, d_c\}$ at each time step $t$, which corresponds to selecting a column from the \ksp matrix. From an implementation standpoint, we zero-fill columns that have not been selected by the agent yet. Hence, the task for the RL agent can also be framed as a mask selection problem where the objective is to obtain a mask $M_T$ that maximizes classification performance. Starting with an initial empty mask $M_0$, the final mask $M_T$ is obtained by iteratively updating the masks as $M_{t-1}[a_t] = 1$. The actions $a_t$ are produced by the policy $\pi_\theta$, which acts on the current state $s_t$. We next describe the state and action spaces.

\begin{figure}[t!]
 \centering
 \includegraphics[width=\linewidth]{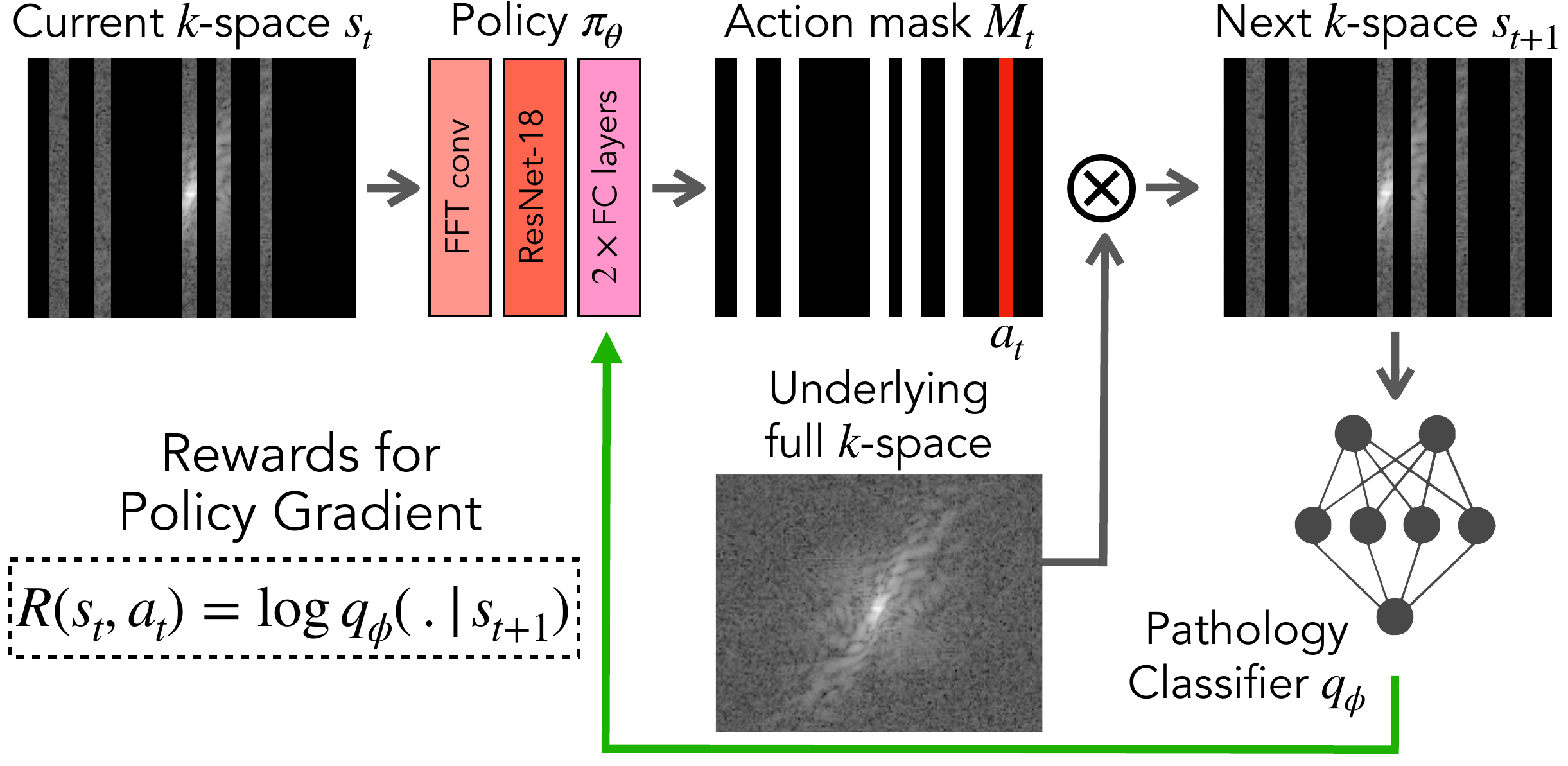}
 \caption{\textbf{\method{} Training.}  \method{} takes an initial sub-sampled \ksp $\mathbf{s}_t$ as input and proposes the next sample $\mathbf{a}_t$ to generate the next state $\mathbf{s}_{t+1}$. This state $\mathbf{s}_{t+1}$ is used by the reward model to compute the log-likelihood $q_\phi(\mathbf{y} \mid \mathbf{x}_{\mathbf{s}_{t+1}})$, the reward for the actor. We repeat these steps for a fixed number of iterations $T$.}
 \label{nn_architecture}
\end{figure}

\paragraph{State space for \method{}:} %
Let $\mathbf{x}^F \in \mathbb{C}^{d_r\times d_c}$ represent the fully acquired \ksp data for a random sample from $\mathcal{D}$. The input state at time $t$ is defined as $s_t= M_{t} \odot \mathbf{x}^F$, where $M_t$ is the mask, a binary vector defined as $M_t \in \prod_{i=1}^{d_c} \{0,1\}$. Initially, at $t=0$ we set $M_0 = 0$ and $s_0 = 0$.
We emphasize that $\mathbf{x}^F$ is utilized merely for notational simplicity. The full k-space data, $\mathbf{x}^F$, is never disclosed to the agent, who can only observe the acquired data. Naturally, during inference, the agent can only observe the \ksp data it has acquired till step $t$. Given an action $a_t$ and state $s_{t+1}$, the next state at time $t+1$ is updated as:
\begin{align}
\label{eqn:state_update}
    s_{t+1} &= M_{t+1} \odot \mathbf{x}^F
\end{align}

\paragraph{Action space for \method{}:}
We use a discrete action space, which is denoted as the set \(\mathcal{A} = \{1, \ldots, d_c\}\) where each element $i$ of $\mathcal{A}$ corresponds to selecting the $i^{\text{th}}$ \ksp column from a given sample. Unlike common RL settings, our action space dynamically shrinks as we iterate through the episode. To handle this shrinkage, we need to iteratively mask out already selected actions, which is discussed next.

\paragraph{Dynamic masking of action space:}
To account for the reduced action space after each selection, we maintain a set $\Omega_{t-1} = \{a_1, \dots, a_{t-1}\}$ to remove actions already selected in previous time steps of the current episode. At the step $t$, we update the set $\Omega_{t} = \Omega_{t-1} \cup \{a_t \}$ and reduce the action space $\mathcal{A}_t = \mathcal{A}_{t-1} \setminus \{\Omega_{t-1}\} $. We generate a mask $M_t$ from the set $\Omega_t$ as:
\begin{align*}
    M_t &= \left( m_{ij} \right)_{i=1}^{d_r},_{j=1}^{d_c} \quad \text{where} \quad m_{ij} = 
\begin{cases} 
1 & \text{if } j \in  \Omega_t \\
0 & \text{otherwise}
\end{cases}
\end{align*}

An alternative approach to address the dynamic masking issue is to assign a negative reward for selecting a previously acquired measurement. However, this is inefficient in practice, and our experiments in \cref{sec:ablation} indicate a strong advantage for our explicit masking technique.  

\subsection{Reward Model}
\label{sec:reward_model}
Typically, RL problem settings have access to an oracle reward model directly as part of the environment, for instance, the game score in Atari \citep{silver2016mastering}. The agent then uses the rewards to optimize the policy parameters $\theta$. In the \method{} setting, we do not have `oracle' rewards but instead have access to underlying pathology labels. However, it is not immediately obvious how to use them to provide intermediate rewards for our agent. 

To address the lack of an oracle reward model, we learn a mapping between the undersampled \ksp data and a quantifiable measure of how well the acquired undersampled \ksp relates to the labels. As such, we repurpose the models used for direct \ksp classification by EMRT \citep{singhal2023feasibility} to compute the log-likelihood between the state $\mathbf{s}_t$ and the label $\mathbf{y}$. 

Since our agent sequentially acquires \ksp data we need our reward model to be able to provide rewards across varying undersampling rates. While a different classifier for each sampling rate can be used, following \citet{jethani2021have, jethani2023don} we train the classifier using random sampling patterns across sampling rates. Such a training approach allows us to train a single classifier that is able to evaluate samples across all rates. For a given state $s_t$ at time $t$, the reward model is defined as 
\begin{equation*}
R(s_t,a_t) = \log q_{\phi}(\mathbf{y}| s_{t+1})
\end{equation*}
where $q_\phi$ is a classifier and $s_{t+1}$ is obtained as in \cref{eqn:state_update}. The classifier is frozen while training the policy.

\paragraph{Policy architecture:}
Following the PPO algorithm \citep{ppo}, we adopt the actor-critic architecture. 
Since our input is in the Fourier domain, we adapt the \textsc{kspace-net} in \citet{singhal2023feasibility} for the actor and critic models. 
The \textsc{kspace-net} first performs a convolution in the Fourier domain \citep{mathieu2013fast}. The output of the convolution, which is complex-valued, is then transformed into real values by taking the magnitude of its inverse Fourier transform. These real-valued features are then fed into a backbone network, where any modern convolutional network such as a ResNet \citep{he2016deep,he2016identity} can be plugged in. In this work, we use a ResNet-18 as the backbone. The features from the backbone are used as input to the actor and critic networks.
The actor-network processes these features into a 2-layer feed-forward layer to produce a categorical distribution over the action space.

\paragraph{Balancing the training environment:}
Traditionally, training environments are sampled uniformly; however, medical datasets can have a highly imbalanced label distribution. For classification problems, this label imbalance can cause the classifier to predict the majority class and still achieve good classification performance. In such situations, a common remedy is to either use importance weighting or up-sampling. In this work, we train the policy by sampling environments in a manner that ensures the agent sees data from all classes with equal probability. For a given dataset labels $\{\mathbf{y}_i\}_{i=1}^N$, we sample an environment $\mathbf{x}$ with probability proportional to the inverse of its label frequency.

\section{Experiments}
\begin{table*}[!ht]
\resizebox{\textwidth}{!}{%

\begin{tabular}{lcccccccc}
                                       & \multicolumn{3}{c}{Knee Dataset}                                      & \multicolumn{4}{c}{Brain Dataset}                                                      & \multicolumn{1}{l}{Prostate Dataset} \\ \hline
\multicolumn{1}{l|}{}                  & ACL           & Mensc. Tear     & \multicolumn{1}{c|}{Abnormal}       & Edema         & Enlg. Ventricles & Mass          & \multicolumn{1}{c|}{Abnormal}       & CS-PCA                               \\ \hline
\multicolumn{1}{l|}{Train slices}      & 29100 (3.6\%) & 29100 (11.55\%) & \multicolumn{1}{c|}{29100 (15.9\%)} & 9420 (1.89\%) & 9420 (1.66\%)    & 9420 (1.83\%) & \multicolumn{1}{c|}{9420 (24.06\%)} & 6649 (5\%)                           \\
\multicolumn{1}{l|}{Validation slices} & 6298 (2.49\%) & 6298 (11.08\%)  & \multicolumn{1}{c|}{6298 (15.39\%)} & 3148 (2.06\%) & 3148 (1.43\%)    & 3148 (2.6\%)  & \multicolumn{1}{c|}{3148 (20.71\%)} & 1431 (4.5\%)                         \\
\multicolumn{1}{l|}{Test slices}       & 6281 (3.58\%) & 6281 (11.94\%)  & \multicolumn{1}{c|}{6281 (16.75\%)} & 3162 (2.44\%) & 3162 (2.3\%)     & 3162 (2.4\%)  & \multicolumn{1}{c|}{3162 (23.88\%)} & 1462 (6\%)                           \\ \hline
\end{tabular}
}
\caption{Number of slices in 
    the training, validation and test splits for each 
    task. The number in the 
    bracket is the percentage of slices with a pathology.}
    \label{table:data}
\end{table*}

Our experiments are designed to answer the following questions:
\begin{itemize}
    \item How does \method{} compare to non-adaptive learned sampling methods?
    \item How do \method{} policies compare to policies optimized for image reconstruction?
    \item How important are the design choices to \method{}'s performance?
\end{itemize}

\begin{figure*}[th]
\centering
    \includegraphics[width=0.9\linewidth]{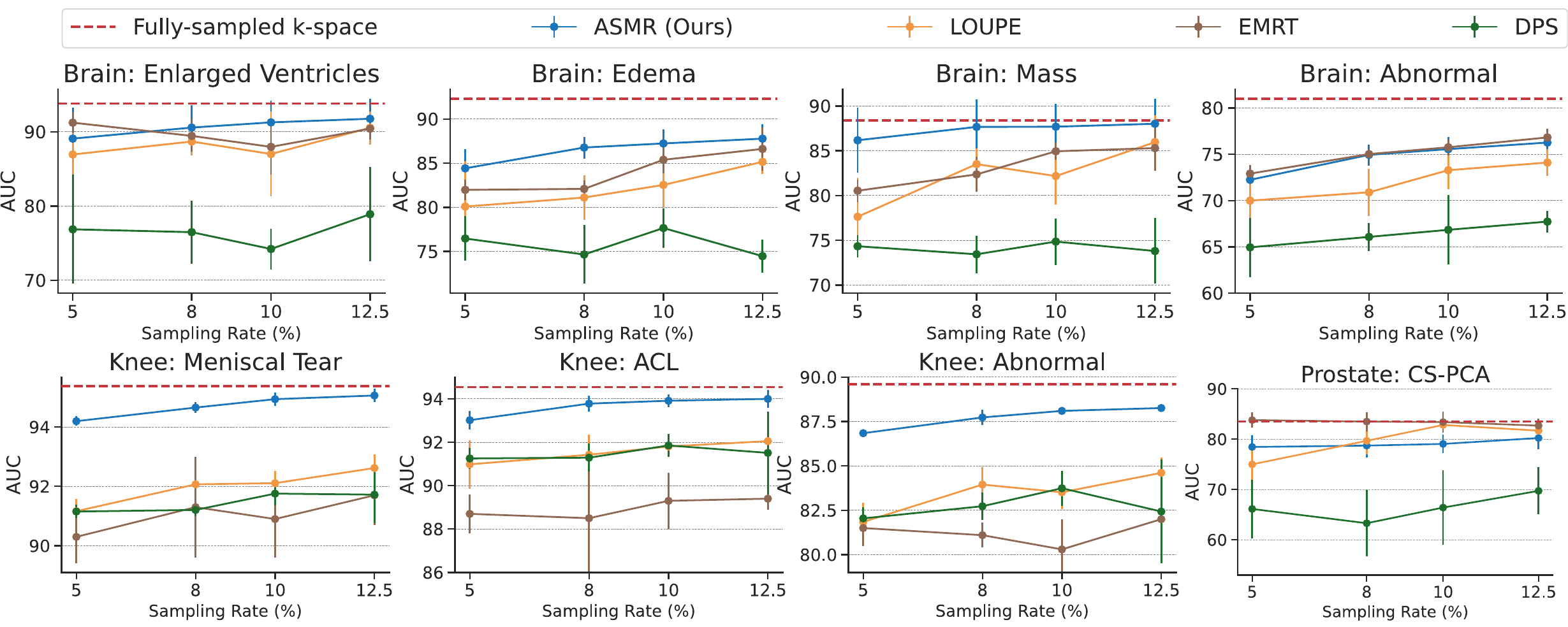}%
  \caption{AUROCs obtained by \method{} compared to Learned Non-Adaptive Methods; the horizontal dotted line in red denotes the performance of an image-based classifier (which uses the entire \ksp data). \method{} outperforms LOUPE and DPS for 6 out of 8 tasks, and outperforms EMRT on 7 out of 8 tasks. All results are computed over 5 seeds, and plotted with their means and standard deviations.}
  \label{fig:learning-based-baseline}
\end{figure*}
\begin{figure*}[ht]
\centering
\includegraphics[width=0.99\textwidth]%
{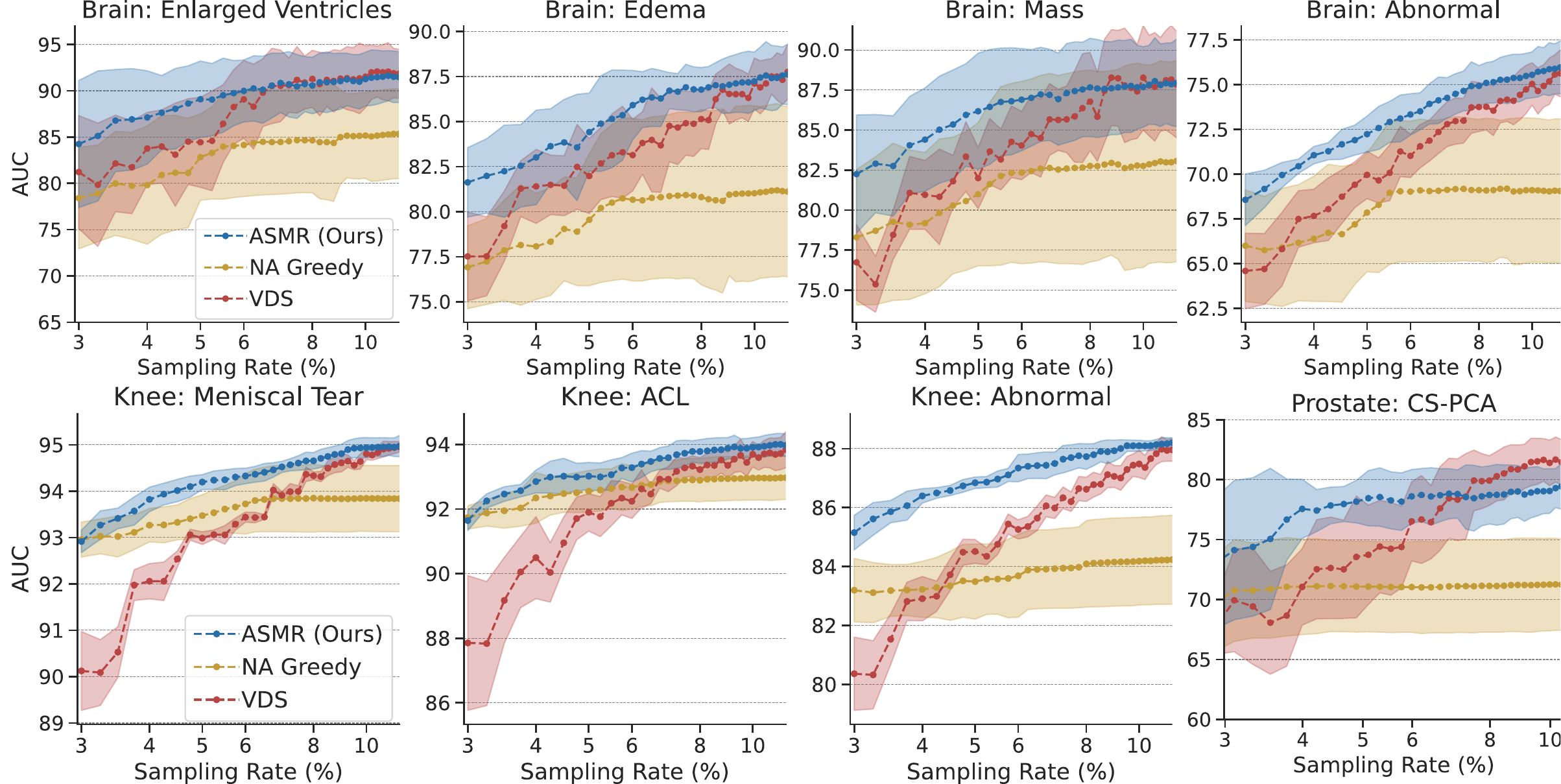}
\caption{AUROCs obtained by \method{} compared to sequential sampling methods such as VDS random and non-adaptive greedy sequence. In the low sampling regime, \method{} consistently outperforms both baselines, while either outperforming or matching their performance at higher sampling rates. All results are computed over 5 seeds, and plotted with their means and standard deviations.}
\label{main_all}
\end{figure*}

\paragraph{Datasets:}
We test \method{} on three different datasets consisting of knee, brain and prostate MR scans:
\begin{itemize}
    \item \textbf{Knee scans}: For the knee dataset, we use the \textsc{FastMRI} knee dataset \citep{zbontar2018fastmri} with slice-level labels provided by \citet{zhao2021fastmri+}. The training, validation, and test splits contain 816, 176, and 175 volumes, respectively. Following \citet{bien2018deep,singhal2023feasibility}, we predict the presence of Meniscal Tears and ACL sprains in each slice. Additionally, we also predict an extra category called abnormal, which is a collection of pathologies that are less frequently observed \citep{bien2018deep} in the dataset.
    
    \item \textbf{Brain scans}: We use the \textsc{FastMRI} brain dataset with labels from \citet{zhao2021fastmri+}.  Our training, validation, and test splits contain 600, 200, and 201 volumes, respectively. We predict the presence of Enlarged Ventricles, Mass, Edema, and an abnormal category.
    
    \item \textbf{Prostate scans}: For the prostate MR scans, we detect the presence of clinically significant prostate cancer (\textsc{CS-PCA}). \textsc{CS-PCA} is defined as a lesion within the prostate for which a radiologist assigns a Prostate Imaging Reporting and Data System (\textsc{PI-RADS}) score \citep{weinreb2016pi} greater or equal to 3. The data was split into 218, 48, and 47 volumes for training, validation, and testing, respectively.     
\end{itemize}
For each dataset, we predict the presence of pathologies in each slice. To prevent label leakage we split each dataset at the volume level, ensuring slices from the same volume do not spill across different splits. Slice level splits along with positivity rates are provided in \cref{table:data}

Due to computational constraints, we use the single-coil \ksp generated from the multi-coil \ksp matrix.  Similar to \citet{zbontar2018fastmri}, we use the emulated single-coil \citep{tygert2020simulating} to convert the multi-coil brain and prostate into single-coil. The emulated single-coil method takes a complex-valued linear combination of coils to produce a single matrix, where the combination of weights is learned per sample. For details see \cite{tygert2020simulating,zbontar2018fastmri}. With additional computional resources, \method{} can be extended to multi-coil by just adapting the \textsc{kspace-net} to accept multi-coil \ksp as input.

\paragraph{Mask Distributions:} In \cref{sec:mask_dist}, we plot the distribution of masks selected by \method{} on each of the datasets.

\paragraph{Evaluation metrics.}
For comparisons on imbalanced datasets, a commonly used metric is the area under the receiver operator curve (\textsc{auroc}). In \cref{sec:appendix_results}, we provide classification metrics other than \textsc{auroc}.

\paragraph{Baselines:} We compare \method{} against several methods, including both adaptive and non-adaptive methods. 
\begin{enumerate}[leftmargin=*]
    \item \textbf{EMRT}: 
    EMRT learns a single classifier and a non-adaptive pattern for each sampling rate.

    \item \textbf{Fully-Sampled Image-based Classifier:} A classifier trained on images generated from the fully-sampled \ksp. This is the standard input for DL models performing image analysis. 
    
    \item \textbf{Learned Non-Adaptive Probabilistic Methods}: For a fixed sampling rate, we repurpose \textsc{loupe} \citep{bahadir2019learning} and \textsc{dps} \citep{huijben2019deep} to learn a classifier and a probabilistic sampling mask jointly. 
    
    \item \textbf{Random Policy}: We use a random policy that samples \ksp columns using the variable density sampling (VDS) prior \citep{lustig2007sparse}.
    
    \item \textbf{Reconstruction Optimized Policy}: We use the greedy policy method in \cite{bakker2020experimental}, using the same 16 channel U-Net \citep{ronneberger2015u} as the reconstruction model using single-coil data.
    
    \item \textbf{Non-Adaptive Greedy Sequence}: A sequential non-adaptive method that uses the same \textsc{kspace-net} used in \method{} to do forward greedy selection \citet{macedo2019theoretical}, producing a sequence of masks $\mathbf{s}_1, \dots, \mathbf{s}_{d}$.

\end{enumerate}

\paragraph{Pathology classifier:} For classifiers that use the \ksp as input we use the \textsc{kspace-net} architecture \citep{singhal2023feasibility} with either a \textsc{ResNet}-18 or \textsc{ResNet}-50 as the backbone architecture. For methods that take an image as input, we use a \textsc{ResNet}-50.

\subsection{Comparing \method{} to non-adaptive methods} 

Following \citep{singhal2023feasibility}, we train EMRT with the \textsc{kspace-net} classifier using random masks sampled from the VDS prior. Then, using the scoring rule defined in \cref{eq:scoring_rule} we sample $K = 100$ masks at a fixed sampling rate from the VDS prior and select the mask that maximizes the score rule. For gradient-based mask learning, we adapt DPS  and LOUPE to learn a probabilistic mask and a classifier jointly. All these methods learn a classifier and a non-adaptive sampling pattern for a fixed sampling rate. 
We train EMRT, DPS, and LOUPE at 5\%, 8\%, 10\%, and 12.5\% sampling rates. Figure \ref{fig:learning-based-baseline} compares the performance of \method{} over these non-adaptive methods. 
As an upper bound, we also provide the performance of an image-based classifier that uses the Inverse Fourier Transform of the fully sampled \ksp as input. We note that \method{} approaches the performance of a fully sampled classifier using just 8\% of the samples.

\subsection{Comparing \method{} to sequential sampling methods}
\method{} sequentially collects \ksp samples, unlike non-adaptive methods like DPS and LOUPE. As such, we compare \method{} to other methods that perform sequential sampling. Following \citet{macedo2019theoretical}, we use the pre-trained classifier $q_\phi$ as in \cref{sec:reward_model} and the scoring rule in \cref{eq:scoring_rule} to iteratively build a sequence $\mathbf{s}_{t+1} = \mathbf{s}_{t} + \arg\max_{j} \mathbf{V}(\mathbf{s}_t + \mathbf{e}_j)$, starting with $\mathbf{s}_0 = 0$. To evaluate the scoring rule, we use a balanced subset of the validation set to build this sequence. This is similar to the greedy non-adaptive oracle considered in \citet{bakker2020experimental}. The greedy sequence suffers from high variance in classification metrics as can be observed across all datasets in Figure~\ref{main_all}, and plateaus beyond a certain sampling rate. 

On average, \method{} achieves 1.87\%, 7.01\%, and 9.82\% absolute gains on AUROC over greedy solutions across the knee, brain, and prostate datasets, respectively.
The VDS random policy, greedy sequence, and \method{} use the same pre-trained classifier for evaluation. Figure~\ref{main_all}
shows a significant performance gap between ASMR and VDS, particularly in the low sampling rate regime ($\leq 8\%$ sampling rate). On average, we see a 1.33\%, 2\%, and 2.25\% gain for \method{} over VDS on the knee, brain, and prostate datasets, respectively.

\subsection{Comparing \method{} to reconstruction optimized policies}
\begin{figure}[ht]
    \centering
    \begin{minipage}{\linewidth}
        \centering
        \includegraphics[width=\linewidth]{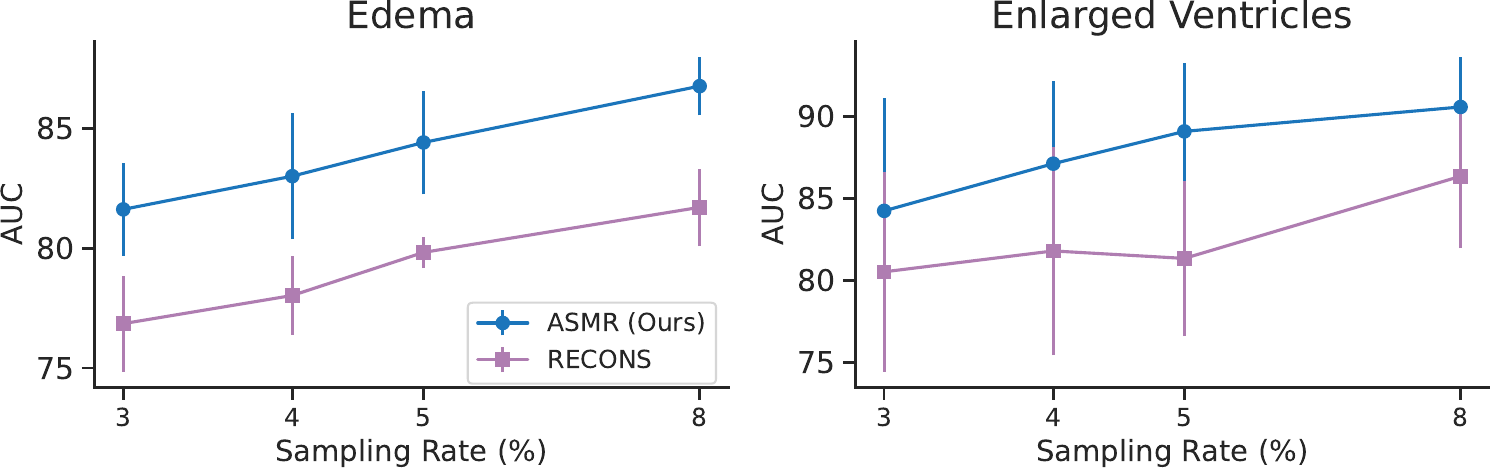} 
    \end{minipage}
    \par
    \begin{minipage}{\linewidth}
        \centering
        \includegraphics[width=\linewidth]{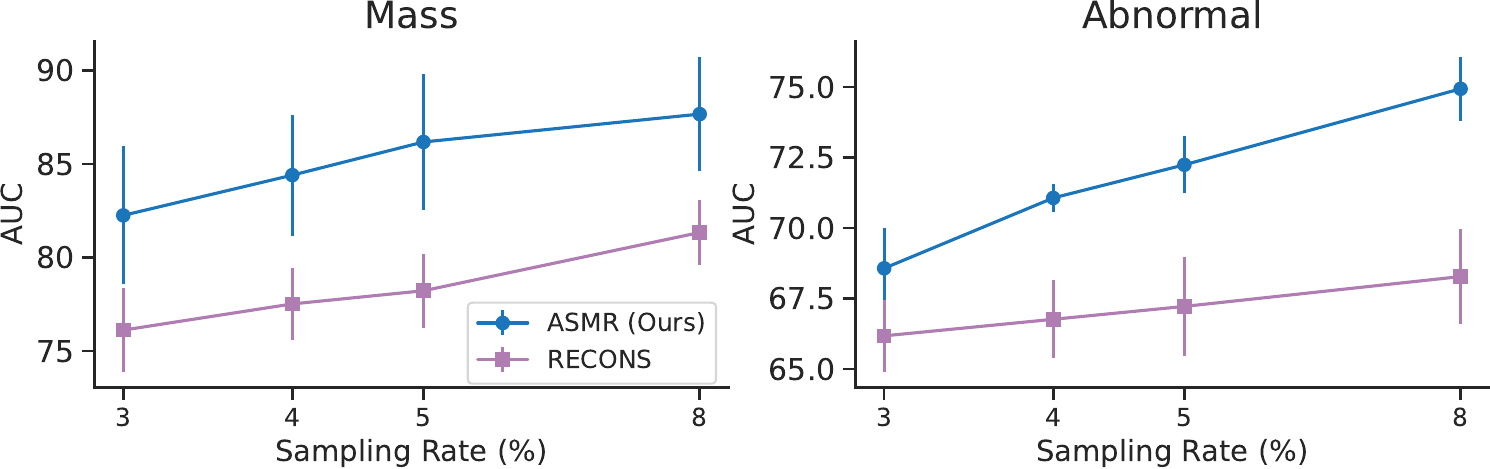} 
        \caption{AUROCs obtained by \method{} compared to a reconstruction-based policy on the Brain test dataset. \method{} significantly outperforms these policies across all sampling rates, indicating the limitations of reconstruction objectives towards pathology prediction.}
        \label{fig:recon_brain}
    \end{minipage}
\end{figure}

To show the benefits of learning policies for classification as opposed to image reconstruction, we follow the open source implementation of \citet{bakker2020experimental} to train a policy model to reconstruct scan images from undersampled \ksp data. Since \citet{bakker2020experimental} requires specifying a target sampling rate, we train separate reconstruction-based policies for each sampling rate. Masks selected by the reconstruction-based policy are then evaluated using the same classifiers used to evaluate \method{}. \cref{fig:recon_brain} compares the two policies in terms of the classification performance of different sampling rates. \method{} consistently outperforms reconstruction-based policies across all four pathologies in the brain dataset, achieving an average of 7\% performance gains. Additional results for knee data and prostate data are included in Appendix~\ref{knee_recon}.

\subsection{Importance of design decisions for \method{}}
\label{sec:ablation}
\begin{figure}[h]
\centering
\includegraphics[width=\linewidth]{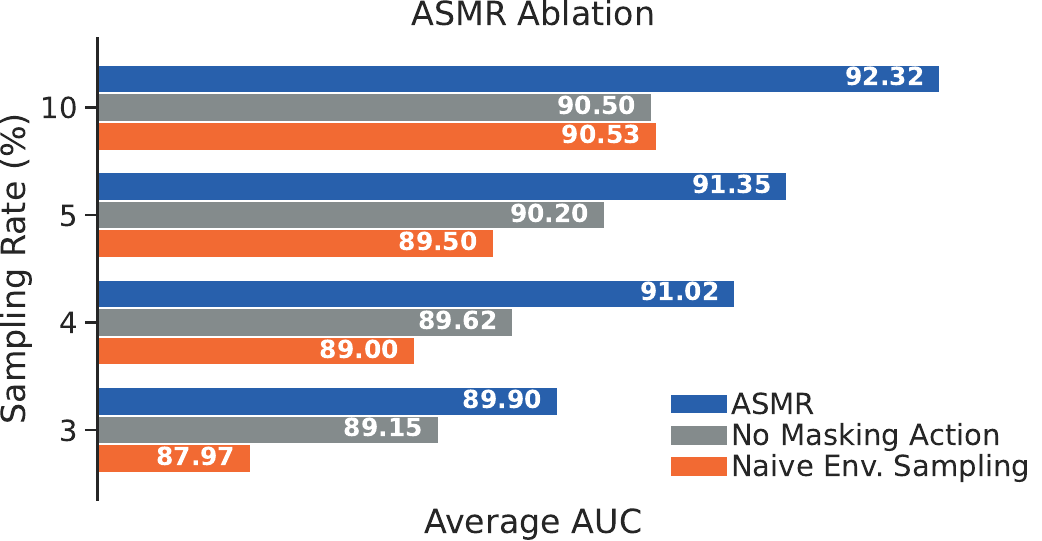}
\caption{Ablation analysis on the Knee dataset for different components of \method{}, where significant performance gain is a achieved.}
\label{fig:ablation}
\end{figure}

To understand the important design decisions of \method{}, we perform two ablations on our training regime choices, using a label-balanced training environment and dynamically masking the action space. We perform these experiments on the knee dataset only and evaluate selected samples by each ablated policy using the same set of classifiers that are used by \method{}. First, we investigate the effect of using a label-balanced training environment. For this, we implement a ``Naive Environment Sampling", that samples the dataset uniformly during the training. Next, we ablate on the dynamic action space masking during training. Instead, we implement a reward penalty of $-1$ when the policy selects a previously acquired \ksp measurement. Figure~\ref{fig:ablation} shows the results of our ablations. We hypothesize that without the label-balanced training environment, the learned policy is biased towards normal patients, causing performance drops across all sampling rates evaluated. Our design choices are crucial to improve the performance of \method{} over a vanilla PPO framework.

\section{Conclusion and Limitations}
\label{sec:conclusion}
We have presented \method{}, an RL-based adaptive sampling technique for pathology prediction. \method{} improves over prior state-of-the-art methods in MR sampling and is a step towards enabling MR-based screening at a population scale as it offers strong pathology prediction using $12 \times$ fewer \ksp samples. However, several challenges still remain before widespread adoption.

\paragraph{Limitations:} \method{} currently only uses single-coil \ksp data while many modern MR scanners now collect multi-coil \ksp data~\cite{deshmane2012parallel}. We believe that \method{} can also be applied to the multi-coil setting as the under-sampling pattern for each coil is the same. Next, we note that due to limited volumetric data, we worked with slice-level classifiers rather than volume-level classifiers. Extending \method{} to volumetric classification or pathology segmentation is an exciting future direction.

\section*{Impact Statement}
Methods that accelerate MR scans and enable wide-spread use of MR technology for screening need to be thoroughly validated prior to clinical adoption. Suddenly increasing the screening population can lead to false positives causing patient trauma and significant cost increases to the healthcare system \citep{kilpelainen2010false,lafata2004economic}. Therefore, prior to clinical adoption, methods such as \method{} need to be improved and undergo extensive clinical trials.

\section*{Acknowledgements}
The computational requirements for this work were supported in part by the resources and personnel of the NYU Langone High Performance Computing (HPC) Core. LP is supported by the Packard Fellowship. We also thank Shenglong Wang and the NYU High Performance Computing for their support.

\nocite{langley00}

\bibliography{references}
\bibliographystyle{icml2024}

\newpage
\appendix
\onecolumn

\section*{Appendix}

\section{Comparison to Learning-based Methods}\label{sec:appendix_results}
Here we present comparisons of \method{} to methods that learn a non-adaptive mask. \cref{knee_loupedps_table} shows the AUC metrics across 3 pathologies in the Knee dataset. We provide the mean and standard deviations across 5 seeds. \cref{brain_loupedps_table} and \cref{prostate_loupedps_table} provide the same results for pathologies in the Brain and Prostate datasets respectively.

\begin{table}[h]
\centering
\begin{tabular}{l||llll}
\hline
Pathology - Sampling Rate (\%) & \multicolumn{1}{c}{DPS} & \multicolumn{1}{c}{EMRT} & \multicolumn{1}{c}{LOUPE} & \multicolumn{1}{c}{\textbf{ASMR (Ours)}}\\ \hline
Abnormal - 5                   & 82.0 ± 0.6 & 81.5 ± 1.0   & 81.8 ± 1.1 & 86.8 ± 0.2           \\
Abnormal - 8                   & 82.7 ± 0.8 & 81.1 ± 0.7 & 83.9 ± 1.0 & 87.7 ± 0.4           \\
Abnormal - 10                  & 83.7 ± 1.0 & 80.3 ± 1.7 & 83.5 ± 0.9 & 88.1 ± 0.2           \\
Abnormal - 12.5                & 82.4 ± 2.9 & 82.0 ± 1.5   & 84.6 ± 0.9 & 88.3 ± 0.2           \\
ACL - 5                        & 91.3 ± 0.5 & 88.7 ± 0.9 & 91.0 ± 1.1 & 93.0 ± 0.4           \\
ACL - 8                        & 91.3 ± 0.7 & 88.5 ± 2.7 & 91.4 ± 0.9 & 93.8 ± 0.4           \\
ACL - 10                       & 91.9 ± 0.5 & 89.3 ± 1.3 & 91.8 ± 0.2 & 93.9 ± 0.3           \\
ACL - 12.5                     & 91.5 ± 1.9 & 89.4 ± 0.5 & 92.1 ± 1.3 & 94.0 ± 0.4           \\
Mensc. Tear - 5                & 91.2 ± 0.2 & 90.3 ± 0.9 & 91.2 ± 0.4 & 94.2 ± 0.2           \\
Mensc. Tear - 8                & 91.2 ± 0.2 & 91.3 ± 1.7 & 92.1 ± 0.6 & 94.7 ± 0.2           \\
Mensc. Tear - 10               & 91.8 ± 0.4 & 90.9 ± 1.3 & 92.1 ± 0.4 & 94.9 ± 0.2           \\
Mensc. Tear - 12.5             & 91.7 ± 1.0 & 91.7 ± 1.0   & 92.6 ± 0.5 & 95.1 ± 0.2           \\ \hline
\end{tabular}
\caption{Test AUC for identifying pathologies under different sampling rates on the Knee dataset}
    \label{knee_loupedps_table}
\end{table}

\begin{table}[!h]
\centering
\begin{tabular}{l||llll}
\hline
Pathology - Sampling Rate (\%) & \multicolumn{1}{c}{DPS} & \multicolumn{1}{c}{EMRT} & \multicolumn{1}{c}{LOUPE} & \multicolumn{1}{c}{\textbf{ASMR (Ours)}} \\ \hline
Edema - 5                      & 76.5 ± 2.5              & 82.0 ± 1.2               & 80.1 ± 5.1                & 84.4 ± 2.2                               \\
Edema - 8                      & 74.7 ± 3.3              & 82.1 ± 0.9               & 81.1 ± 2.6                & 86.8 ± 1.2                               \\
Edema - 10                     & 77.7 ± 2.3              & 85.4 ± 1.5               & 82.5 ± 2.5                & 87.2 ± 1.6                               \\
Edema - 12.5                   & 74.5 ± 1.9              & 86.6 ± 2.5               & 85.1 ± 1.4                & 87.8 ± 1.6                               \\
Enlg. Ventricles - 5           & 76.9 ± 7.3              & 91.2 ± 1.4               & 86.9 ± 3.0                & 89.1 ± 4.1                               \\
Enlg. Ventricles - 8           & 76.5 ± 4.2              & 89.4 ± 2.1               & 88.7 ± 1.9                & 90.6 ± 3.1                               \\
Enlg. Ventricles - 10          & 74.2 ± 2.7              & 88.0 ± 3.7               & 87.0 ± 5.7                & 91.3 ± 2.9                               \\
Enlg. Ventricles - 12.5        & 78.9 ± 6.4              & 90.4 ± 1.5               & 90.5 ± 2.2                & 91.8 ± 2.7                               \\
Mass - 5                       & 74.3 ± 1.2              & 80.5 ± 1.3               & 77.6 ± 4.4                & 86.2 ± 3.6                               \\
Mass - 8                       & 73.4 ± 2.1              & 82.4 ± 2.0               & 83.5 ± 1.7                & 87.7 ± 3.1                               \\
Mass - 10                      & 74.9 ± 2.6              & 84.9 ± 1.0               & 82.2 ± 3.1                & 87.7 ± 2.5                               \\
Mass - 12.5                    & 73.8 ± 3.7              & 85.3 ± 2.5               & 86.0 ± 3.0                & 88.0 ± 2.8                               \\
Abnormal - 5                   & 64.9 ± 3.2              & 72.9 ± 0.9               & 70.0 ± 2.2                & 72.2 ± 1.0                               \\
Abnormal - 8                   & 66.1 ± 1.5              & 75.0 ± 0.4               & 70.9 ± 2.5                & 74.9 ± 1.1                               \\
Abnormal- 10                   & 66.8 ± 3.8              & 75.8 ± 0.9               & 73.3 ± 2.0                & 75.6 ± 1.3                               \\
Abnormal - 12.5                & 67.7 ± 1.2              & 76.8 ± 0.9               & 74.1 ± 1.4                & 76.3 ± 1.5                               \\ \hline
\end{tabular}
\caption{Test AUC for identifying pathologies under different sampling rates on the Brain dataset}
    \label{brain_loupedps_table}
\end{table}

\begin{table}[!htbp]
\centering
\begin{tabular}{l||llll}
\hline
Pathology - Sampling Rate (\%) & \multicolumn{1}{c}{EMRT} & \multicolumn{1}{c}{DPS} & \multicolumn{1}{c}{LOUPE} & \multicolumn{1}{c}{\textbf{ASMR (Ours)}} \\ \hline
CS-PCA - 5                     & 83.8 ± 1.5              & 66.14 ± 5.8              & 75.00 ± 4.1               & 78.46 ± 2.3                              \\
CS-PCA - 8                     & 83.5 ± 1.9              & 63.30 ± 6.6              & 79.68 ± 2.7               & 78.73 ± 2.4                              \\
CS-PCA - 10                    & 83.4 ± 2.1              & 66.41 ± 7.4              & 82.83 ± 0.6               & 79.06 ± 1.9                              \\
CS-PCA - 12.5                  & 82.7 ± 1.4              & 69.73 ± 4.7              & 81.70 ± 1.4               & 80.24 ± 2.2                              \\ \hline
\end{tabular}
\caption{Test AUCs for identifying pathologies under different sampling rates on the Prostate dataset}
    \label{prostate_loupedps_table}
\end{table}

\section{Additional Experiments}
\subsection{Comparison to Reconstruction Optimized
Policies on Knee and Prostate Dataset}
To demonstrate the utility of optimizing a policy for classification performance, following \citet{bakker2020experimental} we also train policies to optimize for reconstruction errors. Using \ksp subsets selected by the reconstruction optimized policy, we evaluate these subsets using our pathology classifier. \cref{knee_recon} shows the results of these experiments and indicate that reconstruction-based objectives are not necessarily suited for classification.
\label{knee_recon}
\begin{figure}[H]
\centering
\includegraphics[width=\textwidth]{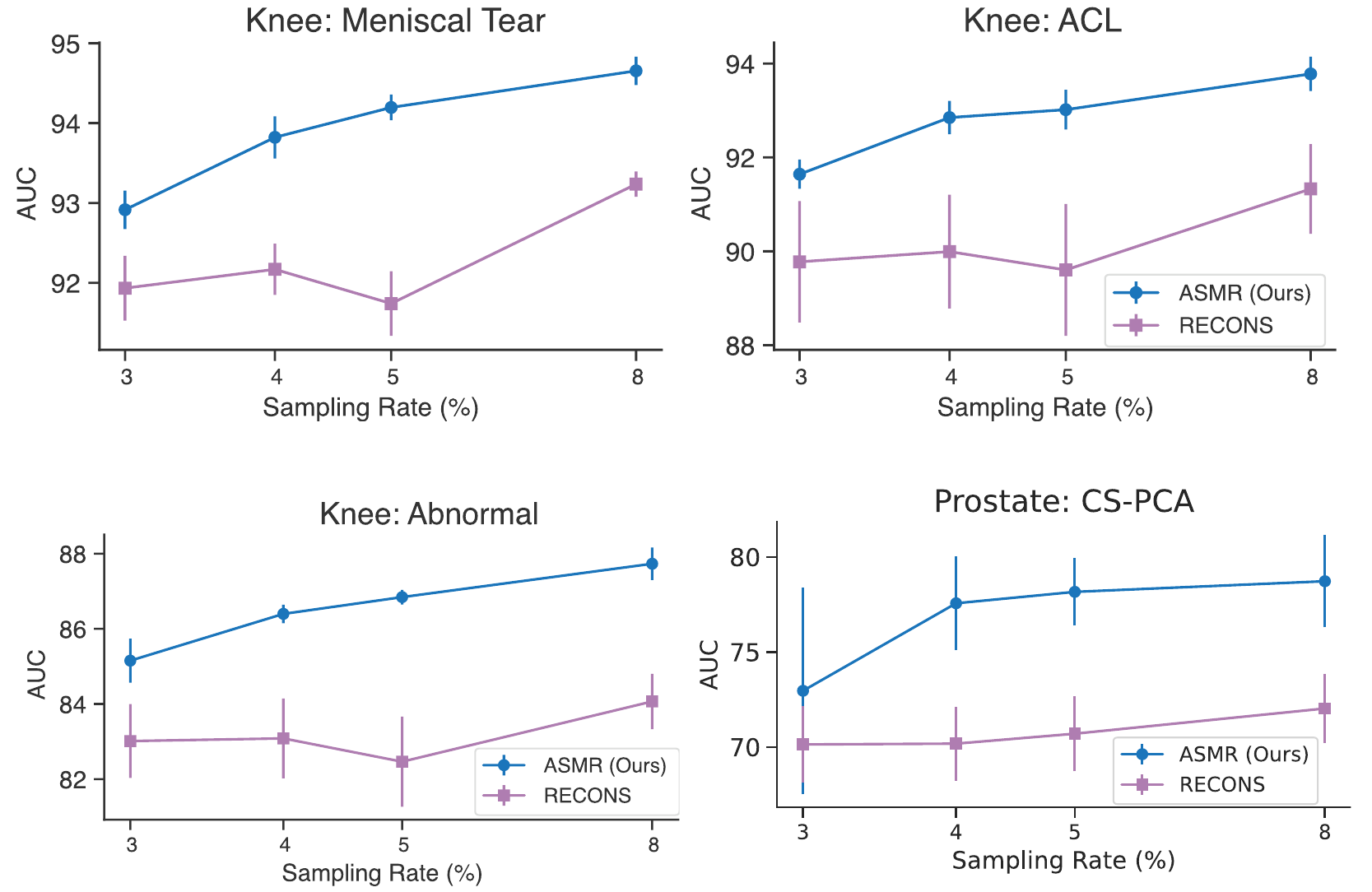}
\end{figure}

\section{Additional Classification Metrics}
\subsection{Knee Results of \method{}}
\begin{table}[h]
\centering
\begin{tabular}{l||cccc}
\hline
Pathology - Sampling Rate (\%) & Balanced Accuracy & Sensitivity & Specifity  & NPV        \\ \hline
ACL - 5                        & 86.1 ± 0.5        & 89.8 ± 0.0  & 82.4 ± 1.1 & 99.5 ± 0.0 \\
ACL - 8                        & 86.5 ± 0.7        & 89.8 ± 0.0  & 83.2 ± 1.5 & 99.5 ± 0.0 \\
ACL - 10                       & 86.8 ± 0.6        & 89.8 ± 0.0  & 83.9 ± 1.2 & 99.6 ± 0.0 \\
ACL - 12.5                     & 86.7 ± 0.3        & 89.8 ± 0.0  & 83.7 ± 0.7 & 99.5 ± 0.0 \\
Mensc. Tear - 5                & 86.8 ± 0.3        & 89.9 ± 0.1  & 83.7 ± 0.5 & 98.4 ± 0.0 \\
Mensc. Tear - 8                & 87.6 ± 0.5        & 89.9 ± 0.0  & 85.3 ± 1.0 & 98.4 ± 0.0 \\
Mensc. Tear - 10               & 88.0 ± 0.3        & 89.9 ± 0.0  & 86.1 ± 0.6 & 98.4 ± 0.0 \\
Mensc. Tear - 12.5             & 88.0 ± 0.5        & 89.9 ± 0.0  & 86.2 ± 1.0 & 98.4 ± 0.0 \\
Abnoraml - 5                   & 76.5 ± 0.9        & 89.9 ± 0.0  & 63.2 ± 1.8 & 96.9 ± 0.1 \\
Abnoraml - 8                   & 77.6 ± 0.8        & 89.9 ± 0.0  & 65.3 ± 1.6 & 97.0 ± 0.1 \\
Abnoraml - 10                  & 77.7 ± 0.4        & 89.9 ± 0.0  & 65.4 ± 0.9 & 97.0 ± 0.0 \\
Abnoraml - 12.5                & 77.6 ± 0.6        & 89.9 ± 0.0  & 65.4 ± 1.2 & 97.0 ± 0.1 \\ \hline
\end{tabular}
\end{table}

\subsection{Brain Results of \method{}}

\begin{table}[h]
\centering
\begin{tabular}{l||cccc}
\hline
Pathology - Sampling Rate (\%) & Balanced Accuracy & Sensitivity & Specifity  & NPV        \\ \hline
Edema - 5                      & 74.2 ± 2.5        & 89.8 ± 0.0  & 82.4 ± 1.1 & 99.5 ± 0.0 \\
Edema - 8                      & 76.9 ± 1.1        & 89.8 ± 0.0  & 83.2 ± 1.5 & 99.5 ± 0.0 \\
Edema - 10                     & 77.6 ± 2.1        & 89.8 ± 0.0  & 83.9 ± 1.2 & 99.6 ± 0.0 \\
Edema - 12.5                   & 78.7 ± 1.9        & 89.8 ± 0.0  & 83.7 ± 0.7 & 99.5 ± 0.0 \\
Enlg. Ventricles - 5           & 78.1 ± 7.1        & 89.9 ± 0.1  & 83.7 ± 0.5 & 98.4 ± 0.0 \\
Enlg. Ventricles - 8           & 78.9 ± 7.2        & 89.9 ± 0.0  & 85.3 ± 1.0 & 98.4 ± 0.0 \\
Enlg. Ventricles - 10          & 79.8 ± 7.3        & 89.9 ± 0.0  & 86.1 ± 0.6 & 98.4 ± 0.0 \\
Enlg. Ventricles - 12.5        & 81.2 ± 7.3        & 89.9 ± 0.0  & 86.2 ± 1.0 & 98.4 ± 0.0 \\
Mass - 5                       & 74.6 ± 3.7        & 89.9 ± 0.0  & 63.2 ± 1.8 & 96.9 ± 0.1 \\
Mass - 8                       & 76.5 ± 3.3        & 89.9 ± 0.0  & 65.3 ± 1.6 & 97.0 ± 0.1 \\
Mass - 10                      & 77.1 ± 2.3        & 89.9 ± 0.0  & 65.4 ± 0.9 & 97.0 ± 0.0 \\
Mass - 12.5                    & 77.0 ± 1.8        & 89.9 ± 0.0  & 65.4 ± 1.2 & 97.0 ± 0.1 \\
Abnormal - 5                   & 62.1 ± 2.0        & 72.9 ± 0.9  & 70.0 ± 2.2 & 72.2 ± 1.0 \\
Abnormal - 8                   & 64.1 ± 1.3        & 75.0 ± 0.4  & 70.9 ± 2.5 & 74.9 ± 1.1 \\
Abnormal- 10                   & 64.9 ± 1.8        & 75.8 ± 0.9  & 73.3 ± 2.0 & 75.6 ± 1.3 \\
Abnormal - 12.5                & 65.4 ± 1.6        & 76.8 ± 0.9  & 74.1 ± 1.4 & 76.3 ± 1.5 \\ \hline
\end{tabular}
\end{table}

\subsection{Prostate Results of \method{}}

\begin{table}[h]
\centering
\begin{tabular}{l||cccc}
\hline
Pathology - Sampling Rate (\%) & Balanced Accuracy & Sensitivity & Specifity  & NPV        \\ \hline
CS-PCA - 5                     & 68.3 ± 2.0        & 89.7 ± 0.0  & 47.0 ± 3.9 & 98.4 ± 0.1 \\
CS-PCA - 8                     & 70.0 ± 3.7        & 89.7 ± 0.0  & 50.2 ± 7.5 & 98.5 ± 0.2 \\
CS-PCA - 10                    & 68.4 ± 3.0        & 89.7 ± 0.0  & 47.2 ± 6.0 & 98.4 ± 0.2 \\
CS-PCA - 12.5                  & 68.0 ± 2.5        & 89.7 ± 0.0  & 46.4 ± 5.1 & 98.4 ± 0.2 \\ \hline
\end{tabular}
\end{table}

\section{Training Details}
\subsection{Training Parameters}
The \ksp data input to agent is in complex domain with height and width of ${d_r\times d_c}$. The ${d_r\times d_c}$ for knee, brain, prostate are $(768\times 400)$, $(640\times 400)$, and $(320\times 451)$, respectively. Hyperparameters used to train the policy are provided in Table \ref{table:hyper_params}.

\begin{table}[!h]
\centering
\begin{tabular}{|l|l|}
\hline
Parameter                  & Value \\ \hline
Optimizer                  & Adamw \\
Learning rate              & 1e-04 \\
weight decay               & 1e-04 \\
discount factor            & 0.99  \\
gae\_lambda                & 0.95  \\
clip ratio        & 0.2   \\
entropy cost               & 0.01  \\
grad norm clipping         & 0.5   \\
value function coefficient & 0.5   \\
parallelized rollout       & 128    \\ \hline
\end{tabular}
\caption{Hyperparameters of our agent}
\label{table:hyper_params}
\end{table}
\newpage
\section{Policy selected Mask Distributions}
\label{sec:mask_dist}
Below we plot a heatmap of the \ksp column selections made by \method{} on the Knee, Brain and Prostate datasets. We evaluate our policy to have a sampling rate of $12.5\%$ and generate heat maps representing the fraction of samples in the test set for which a column was picked.

\begin{figure}[h]
\centering
\includegraphics[width=0.43\linewidth]{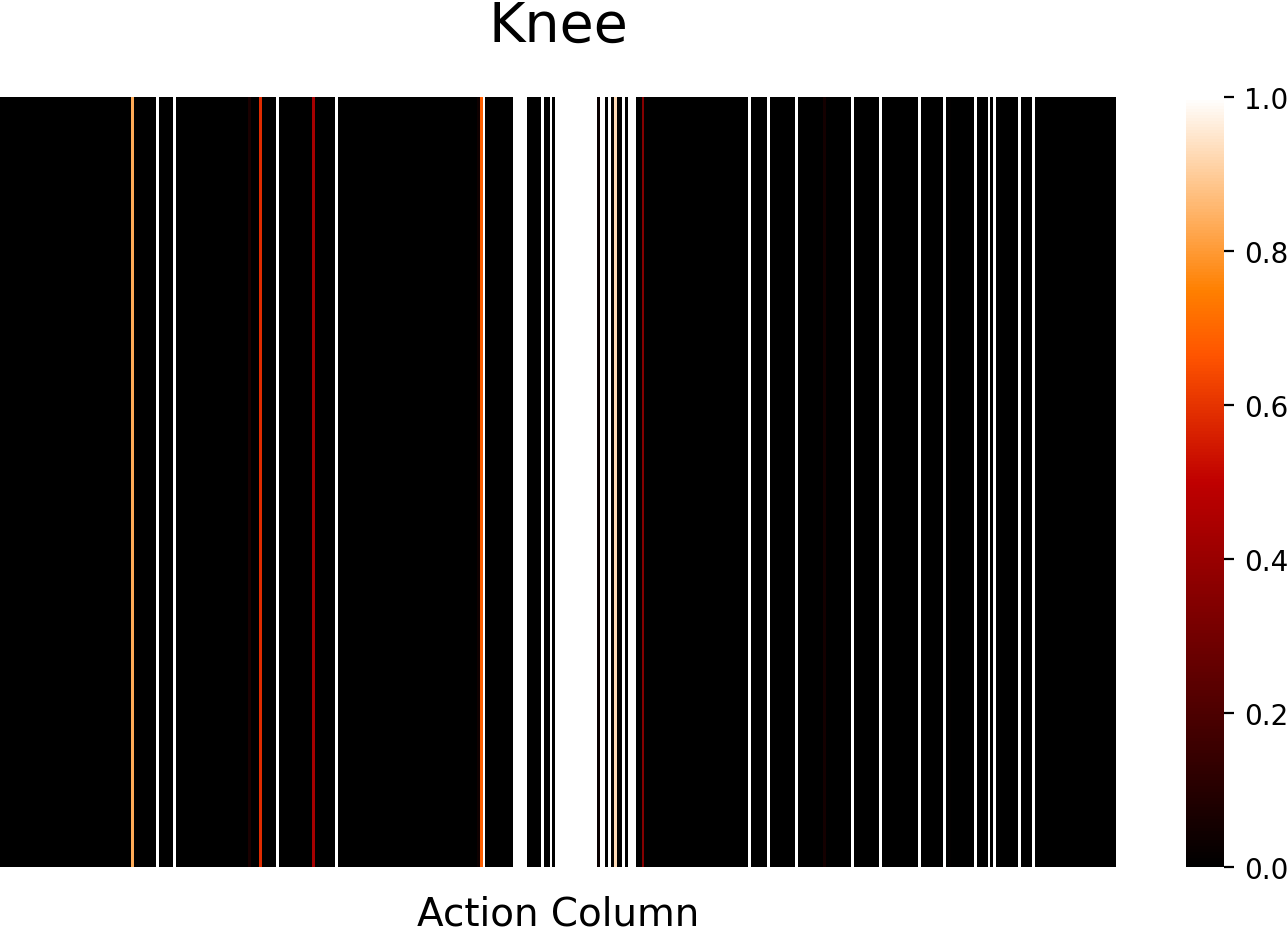}
\end{figure}

\begin{figure}[h]
\centering
\includegraphics[width=0.43\linewidth]{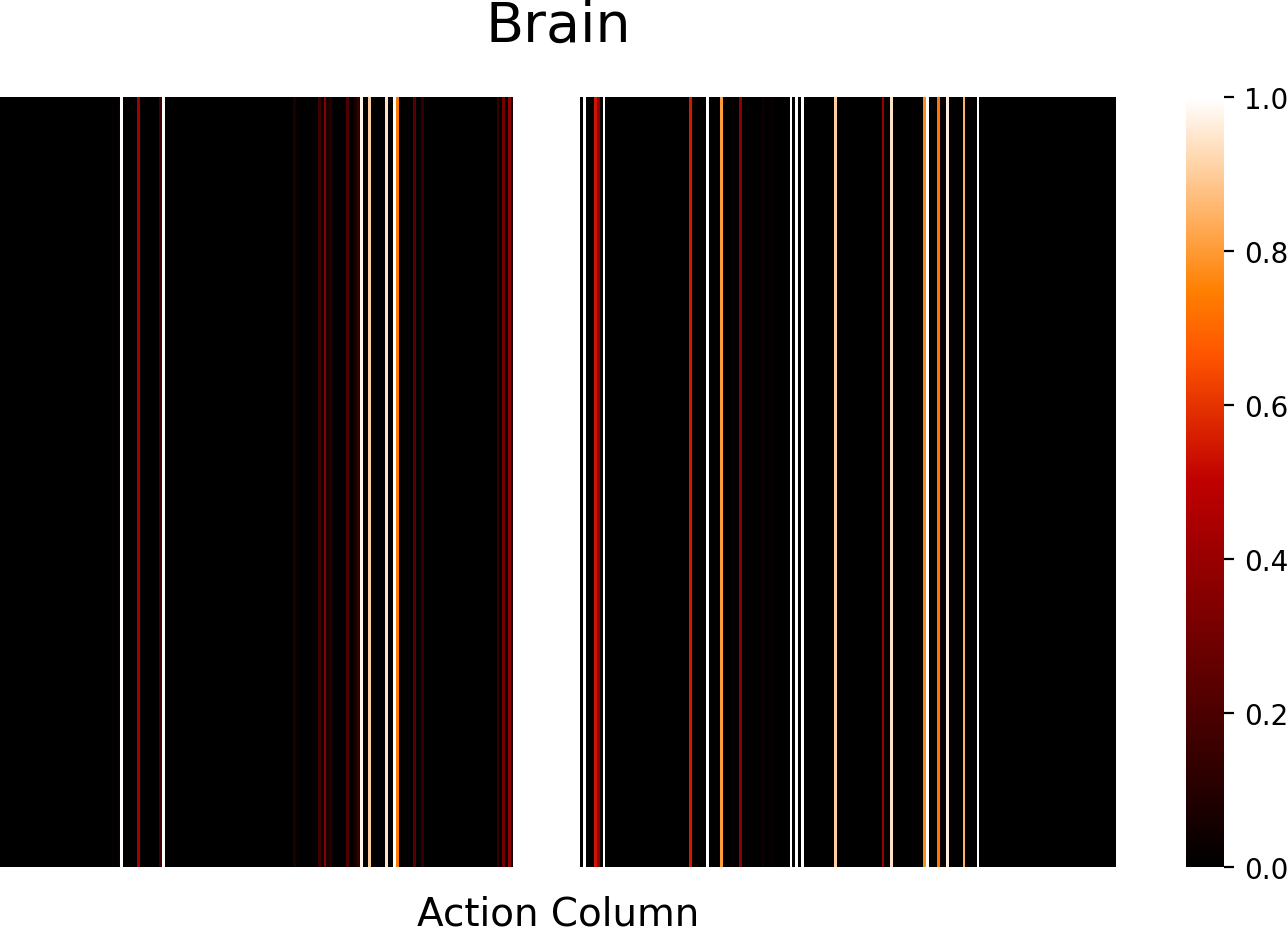}
\end{figure}

\begin{figure}[h]
\centering
\includegraphics[width=0.43\linewidth]{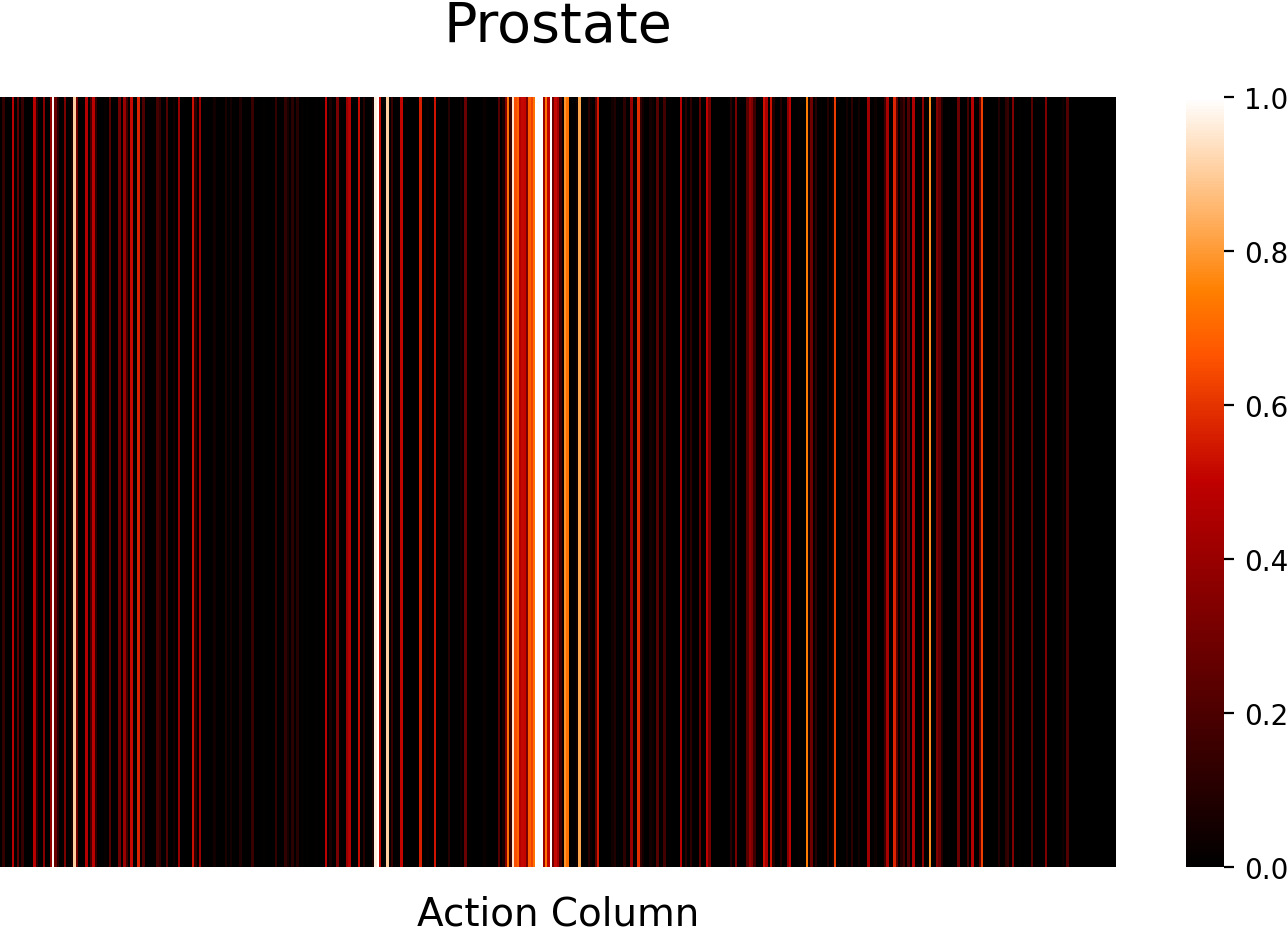}
\end{figure}

\onecolumn

\end{document}